\documentclass[runningheads]{llncs}

\usepackage[T1]{fontenc}
\usepackage[utf8]{inputenc}
\usepackage{graphicx}
\usepackage{amsmath,amssymb,amsfonts}
\usepackage{algorithmic}
\usepackage{textcomp}
\usepackage{xcolor}
\usepackage{subfig}
\usepackage{gensymb}
\usepackage[numbers]{natbib}
\usepackage{svg}
\usepackage{mwe}
\usepackage{verbatim}
\usepackage{amsbsy}
\usepackage{siunitx}
\usepackage{tabularx, booktabs}
\usepackage{soul}
\usepackage{tikz}
\usetikzlibrary{calc}
\usepackage{float}
\usepackage{hyperref}
\hypersetup{colorlinks=true, linkcolor=blue, citecolor=blue, urlcolor=blue}

\DeclareUnicodeCharacter{0456}{\i} 

\begin{document}

\title{UruBots Autonomous Cars Challenge Pro Team Description Paper for FIRA 2025}

\author{Pablo Moraes\inst{1} \and Mónica Rodríguez\inst{1} \and Sebastian Barcelona\inst{1} \and Angel Da Silva\inst{1} \and Santiago Fernandez\inst{1} \and Hiago Sodre\inst{1} \and Igor Nunes\inst{1} \and Bruna Guterres\inst{1} \and Ricardo Grando\inst{1}}
\institute{Technological University of Uruguay}  


\authorrunning{UruBots AC Pro et al.}
\titlerunning{UruBots AC Pro}

\maketitle  

\begin{abstract}

This paper describes the development of an autonomous car by the UruBots team for the 2025 FIRA Autonomous Cars Challenge (Pro). The project involves constructing a compact electric vehicle, approximately the size of an RC car, capable of autonomous navigation through different tracks. The design incorporates mechanical and electronic components and machine learning algorithms that enable the vehicle to make real-time navigation decisions based on visual input from a camera. We use deep learning models to process camera images and control vehicle movements. Using a dataset of over ten thousand images, we trained a Convolutional Neural Network (CNN) to drive the vehicle effectively, through two outputs, steering and throttle. The car completed the track in under 30 seconds, achieving a pace of approximately 0.4 meters per second while avoiding obstacles.

\end{abstract}

\section{Introduction}

The field of autonomous vehicles is continuously evolving, driving new technological solutions in small-scale transportation systems \cite{pehlivan2020real}. Competitions such as the FIRA Autonomous Cars Challenge contribute to this progress by encouraging the practical application of emerging technologies and promoting innovation among teams worldwide. For the 2025 challenge edition, our team has developed a fully autonomous, RC-sized electric vehicle designed to navigate dynamic tracks with precision and reliability.

The system integrates mechanical and electronic components in the hardware, combined with deep learning techniques in the software. Using data collected by a front-facing camera, we trained models that allow the vehicle to interpret its environment, follow a predefined path, and effectively avoid obstacles. This approach builds upon previous work on behavior cloning for autonomous driving \cite{moraes2024behavior} and the advancements described in our team's technical paper from last year's competition \cite{moraes2024urubots}.

This work presents the components of our project, detailing both the physical setup and the control architecture. In terms of hardware, we describe the mechanical layout, the selection of electronic parts, and how the sensors were integrated into the system. On the software side, we explain the algorithms used, how the data is processed, and the logic behind the vehicle’s autonomous behavior. The document concludes with a summary of test results obtained on a track designed for evaluation.

\subsection{Hardware}

\begin{figure}[H]
    \centering
    \includegraphics[width=0.8\linewidth]{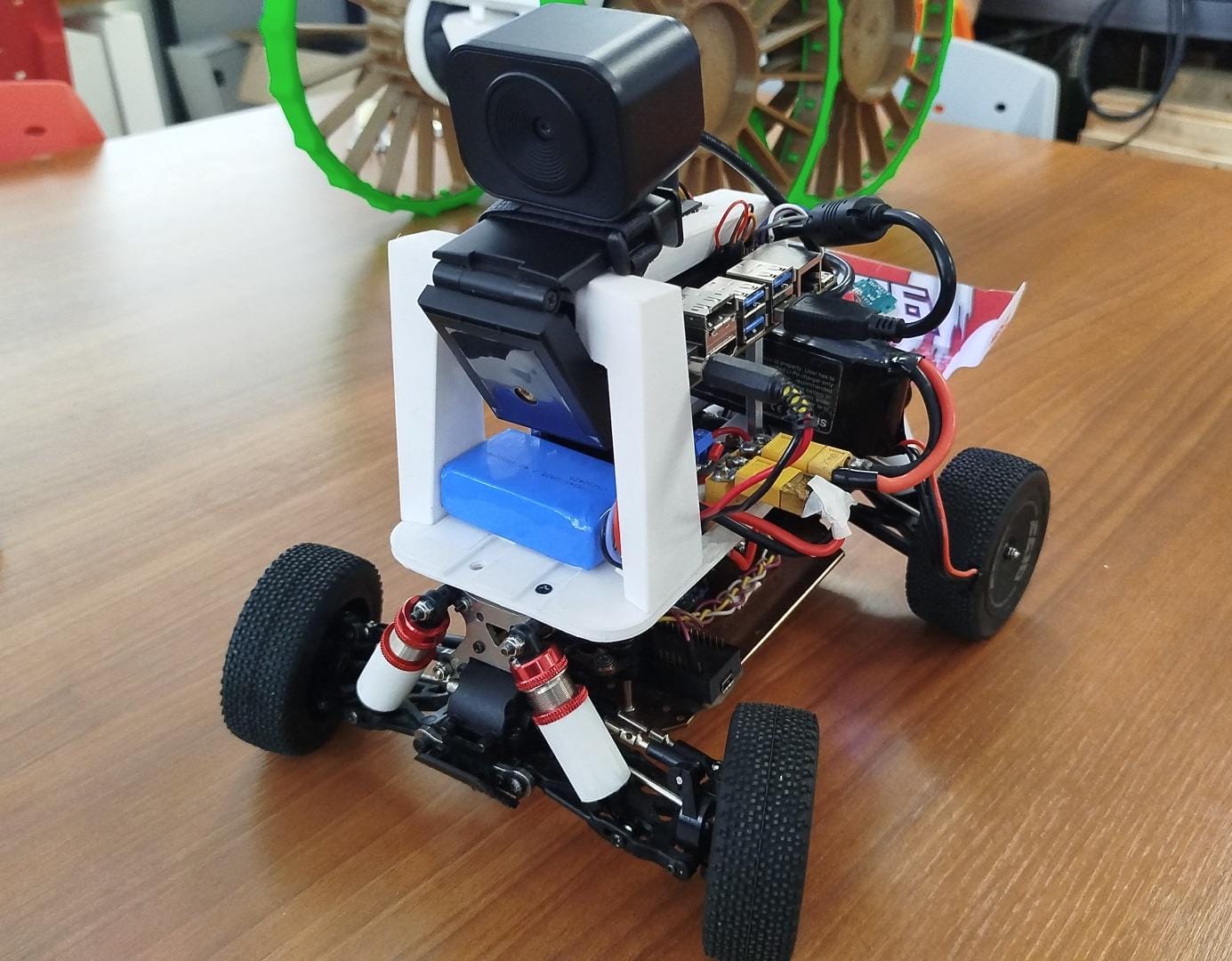}
    \caption{The autonomous car during the construction phase.}
    \label{fig:car}
\end{figure}

For the Autonomous Cars PRO competition, our team started with the chassis of an RC car model, WLtoys 144001, and redesigned it to develop a fully autonomous vehicle, getting as result the car shown in \ref{fig:car}. Our team modified the original structure to improve stability and precision, while providing a solid base for integrating all the necessary hardware and software components for autonomous operation.

The mechanical design focused on maximizing the stability of the vision system by removing the original shock absorbers and replacing them with rigid 3D-printed structures to fix the suspension and avoid angle variations in the camera during movement.

The new locomotion system includes a waterproof brushed DSPower 540 45T motor powered and controlled by a 1060 RTR 60A brushed ESC. This configuration was chosen for its improved control at low speeds, critical for the second stage of the FIRA challenge where precise forward and reverse movements are required.

\begin{figure}[H]
  \centering
  \subfloat[MPU9250 IMU.\label{fig:imu}]{
    \begin{minipage}[c][\width]{0.14\textwidth}
       \centering
       \includegraphics[width=\textwidth]{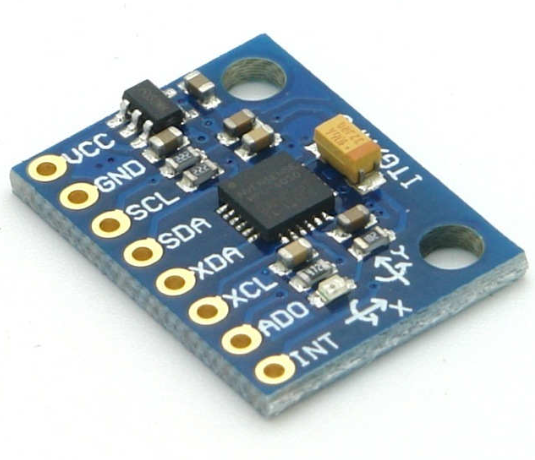}
    \end{minipage}}
  \hfill     
  \subfloat[Jetson Nano 4GB.\label{fig:jetson}]{
    \begin{minipage}[c][\width]{0.18\textwidth}
       \centering
       \includegraphics[width=\textwidth]{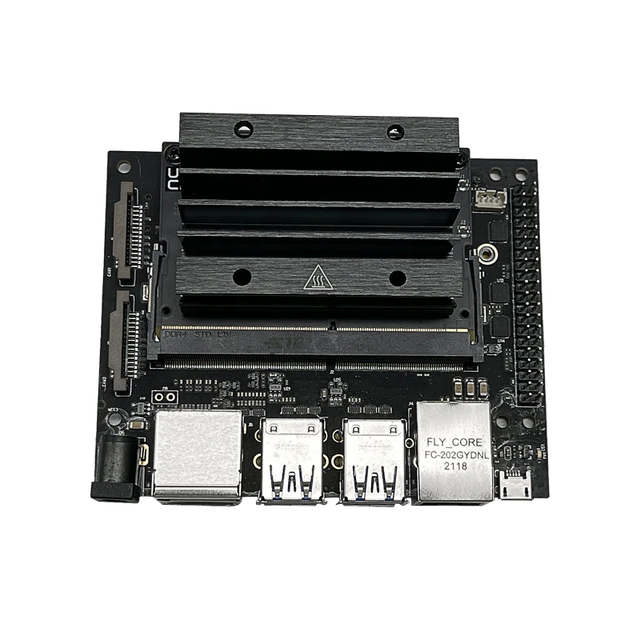}
    \end{minipage}}
  \hfill     
  \subfloat[k4 Camera.\label{fig:camera}]{
    \begin{minipage}[c][\width]{0.14\textwidth}
       \centering
       \includegraphics[width=\textwidth]{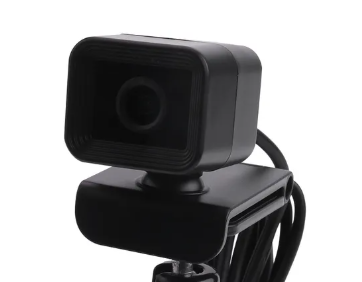}
    \end{minipage}}
  \hfill     
  \subfloat[Two 11.1V 3S LiPo Batteries.\label{fig:battery}]{
    \begin{minipage}[c][\width]{0.16\textwidth}
       \centering
       \includegraphics[width=\textwidth]{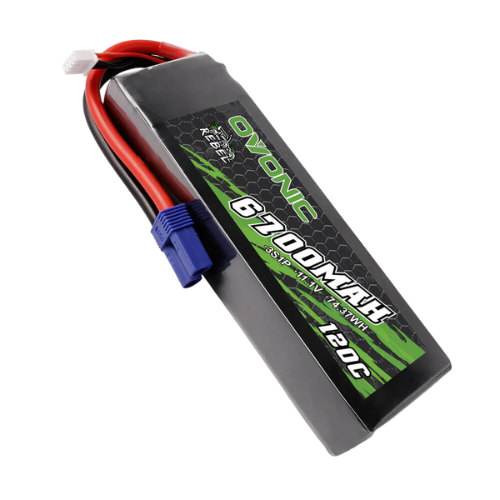}
    \end{minipage}}
  \hfill     
  \subfloat[Brushed Motor DSPower 540 45T.\label{fig:motors}]{
    \begin{minipage}[c][\width]{0.20\textwidth}
       \centering
       \includegraphics[width=\textwidth]{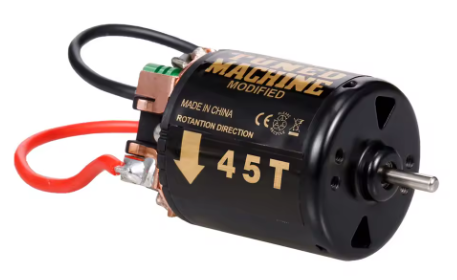}
    \end{minipage}}
  \hfill     
  \subfloat[Servomotor PDI-1109MG.\label{fig:servo}]{
    \begin{minipage}[c][\width]{0.18\textwidth}
       \centering
       \includegraphics[width=\textwidth]{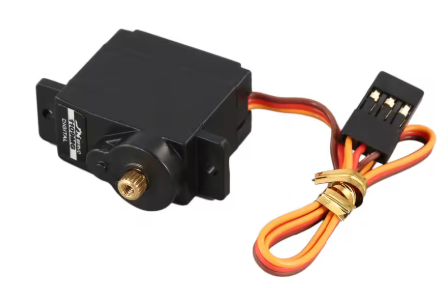}
    \end{minipage}}
  \hfill     
  \subfloat[PCA9685 PWM Controller.\label{fig:control}]{
    \begin{minipage}[c][\width]{0.18\textwidth}
       \centering
       \includegraphics[width=\textwidth]{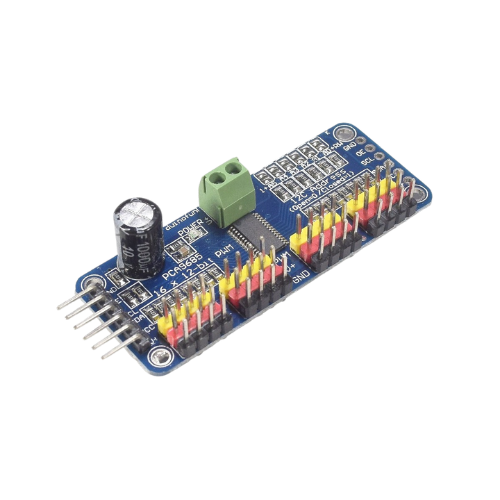}
    \end{minipage}}
  \hfill
  \subfloat[Brushed ESC (Motor Driver).\label{fig:esc}]{
    \begin{minipage}[c][\width]{0.18\textwidth}
       \centering
       \includegraphics[width=\textwidth]{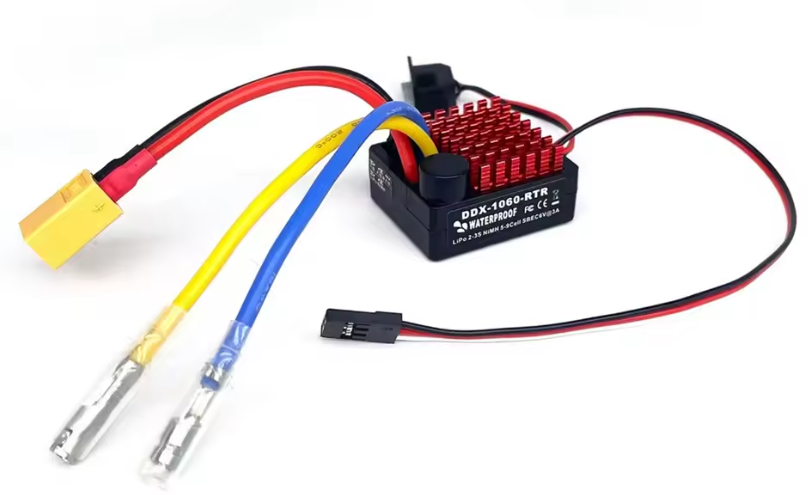}
    \end{minipage}}
  \hfill
  \subfloat[WiFi-Bluetooth Driver Board.\label{fig:wifi}]{
    \begin{minipage}[c][\width]{0.12\textwidth}
       \centering
       \includegraphics[width=\textwidth]{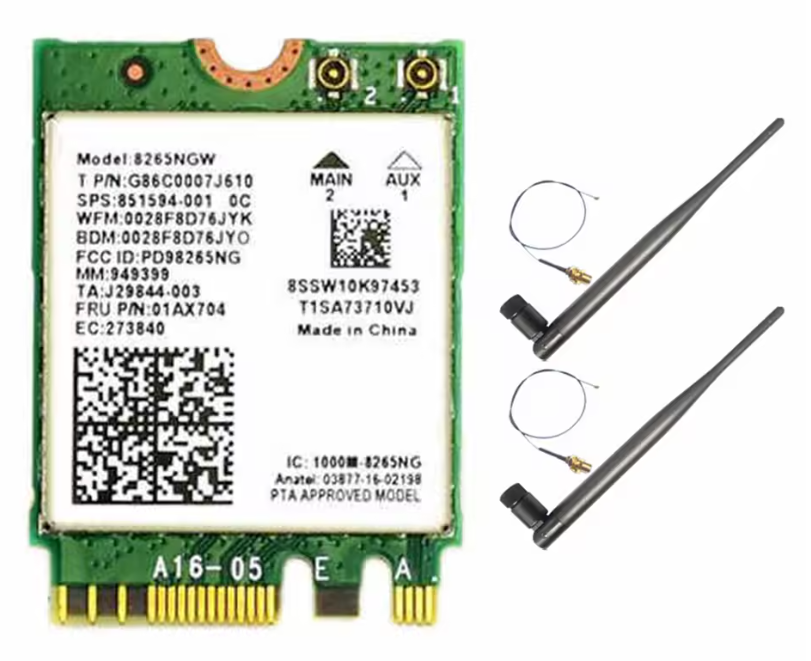}
    \end{minipage}}
  \hfill     
  \subfloat[Step-Down Converter for Jetson Nano Power Supply.\label{fig:stepdown_jetson}]{
    \begin{minipage}[c][\width]{0.18\textwidth}
       \centering
       \includegraphics[width=\textwidth]{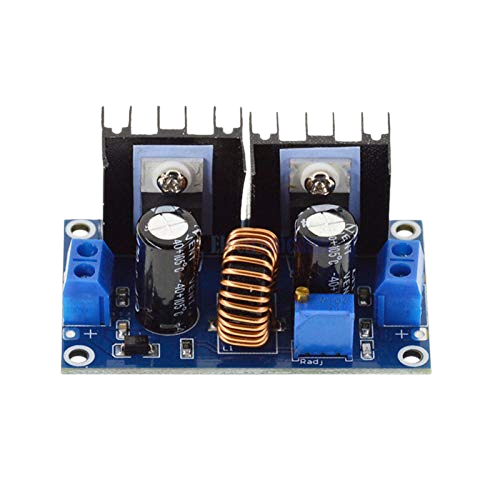}
    \end{minipage}}
  \hfill
  \subfloat[Step-Down Converter for Servo Power Supply (PCA9685 Board).\label{fig:stepdown_servo}]{
    \begin{minipage}[c][\width]{0.18\textwidth}
       \centering
       \includegraphics[width=\textwidth]{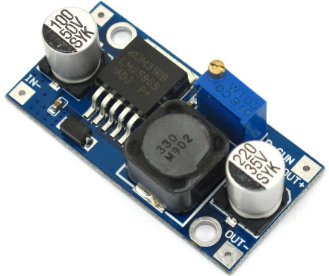}
    \end{minipage}}
  \hfill
  \subfloat[Optical Encoder.\label{fig:encoder}]{
    \begin{minipage}[c][\width]{0.18\textwidth}
       \centering
       \includegraphics[width=\textwidth]{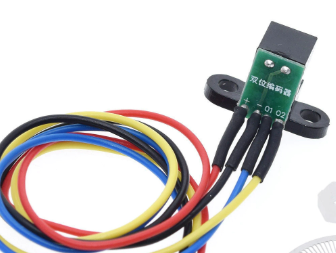}
    \end{minipage}}
  \hfill
  \subfloat[WLtoys 144001 RC Car Chassis.\label{fig:chassis}]{
    \begin{minipage}[c][\width]{0.12\textwidth}
       \centering
       \includegraphics[width=\textwidth]{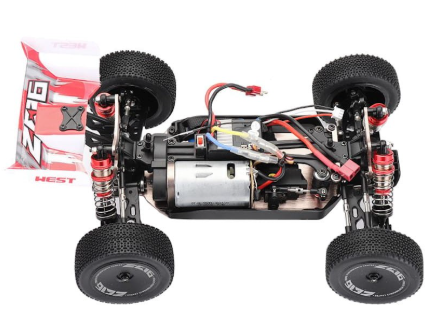}
    \end{minipage}}
  \hfill     
  \subfloat[Arduino Nano.\label{fig:arduino}]{
    \begin{minipage}[c][\width]{0.18\textwidth}
       \centering
       \includegraphics[width=\textwidth]{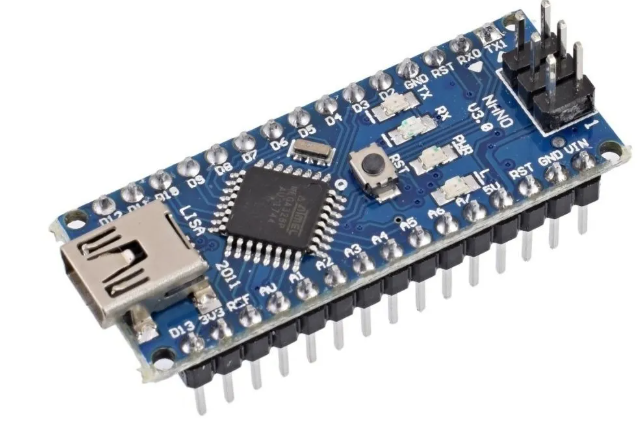}
    \end{minipage}}
  \caption{Components of our autonomous vehicle.}
  \label{componentes}
\end{figure}

The central control system is based on a Jetson Nano 4GB, which processes images in real-time from a USB-connected camera. Speed and steering commands are managed by a PCA9685 PWM controller connected via I2C to the Jetson Nano. The PCA9685 generates PWM signals to both the ESC and the steering servo motor.

The power system consists of two separate 3S LiPo batteries. One battery feeds a step-down converter that regulates the voltage to a stable 5V supply for the Jetson Nano, ensuring it operates at full performance. The other battery powers the ESC directly for the brushed motor and also supplies a separate step-down converter that provides 5V to the PCA9685 board, which in turn powers the steering servo motor.

An Arduino Nano connected via USB serial to the Jetson Nano reads sensor data from an MPU9250 inertial measurement unit (IMU) and an optical encoder mounted on the driveshaft (the main axis of the 4x4 vehicle). The Arduino processes these sensor signals locally and sends the data to the Jetson Nano, enabling accurate speed and orientation tracking.
The entire system is mounted on the WLtoys 144001 chassis using custom-designed 3D-printed mounts specifically created for this project.

In the Table \ref{tab:car_components} all hardware components and their roles are described, as well as their aspects, shown in Figure \ref{componentes}.

\begin{table}[H]
    \centering
    \caption{Main hardware components and their roles in the autonomous vehicle.}
    \label{tab:car_components}
    \renewcommand{\arraystretch}{1.2}
    \begin{tabularx}{\linewidth}{|p{4cm}|X|}
    \hline
    \textbf{Component} & \textbf{Description} \\
    \hline
    Jetson Nano 4GB (Fig.~\ref{fig:jetson}) & Main processing unit that runs the AI models and coordinates control signals. Communicates via USB and I2C. \\
    \hline
    USB Camera (K4) (Fig.~\ref{fig:camera}) & Captures real-time front view images used for navigation decisions. \\
    \hline
    Arduino Nano (Fig.~\ref{fig:arduino}) & Interfaces with IMU and encoder; transmits processed sensor data to the Jetson Nano. \\
    \hline
    MPU9250 IMU (Fig.~\ref{fig:imu}) & Detects orientation and acceleration, assisting in estimating the car’s motion state. \\
    \hline
    Optical Encoder (Fig.~\ref{fig:encoder}) & Measures rotational velocity and direction of the drivetrain. \\
    \hline
    PCA9685 PWM Controller (Fig.~\ref{fig:control}) & Generates PWM signals to control steering and throttle based on Jetson Nano commands. \\
    \hline
    Servo Motor PDI-1109MG (Fig.~\ref{fig:servo}) & Adjusts the front wheel angle based on control input for navigation. \\
    \hline
    Brushed ESC 1060 RTR 60A (Fig.~\ref{fig:esc}) & Controls motor speed and direction; receives PWM input from the PCA9685. \\
    \hline
    DSPower 540 45T Motor (Fig.~\ref{fig:motors}) & Provides propulsion for the vehicle; selected for precise low-speed control. \\
    \hline
    3S LiPo Batteries (x2) (Fig.~\ref{fig:battery}) & Independent power sources for the computation system and motor system. \\
    \hline
    Step-Down Converter (Jetson) (Fig.~\ref{fig:stepdown_jetson}) & Regulates LiPo voltage down to 5V for stable Jetson Nano operation. \\
    \hline
    Step-Down Converter (PCA9685) (Fig.~\ref{fig:stepdown_servo}) & Regulates power for the servo control board and steering motor. \\
    \hline
    WLtoys 144001 Chassis (Fig.~\ref{fig:chassis}) & Mechanical base platform, modified for rigidity and component mounting. \\
    \hline
    WiFi-Bluetooth Driver Board (Fig.~\ref{fig:wifi}) & Enables wireless communication and potential remote debugging or telemetry. \\
    \hline
    \end{tabularx}
\end{table}

\subsection{Software}

The software architecture of our autonomous vehicle builds upon the Donkey Car open-source framework, a platform designed for developing self-driving robotic cars with Python that are adaptable to boards like Jetson Nano and Raspberry Pi. Our implementation leverages this framework to handle the vehicle’s steering and throttle control using real-time data processed on a Jetson Nano. The primary sensor input consists of images captured by a monocular USB camera facing forward, providing continuous visual feedback from the track.

\begin{figure}[H]
    \centering
    \includegraphics[scale=0.3]{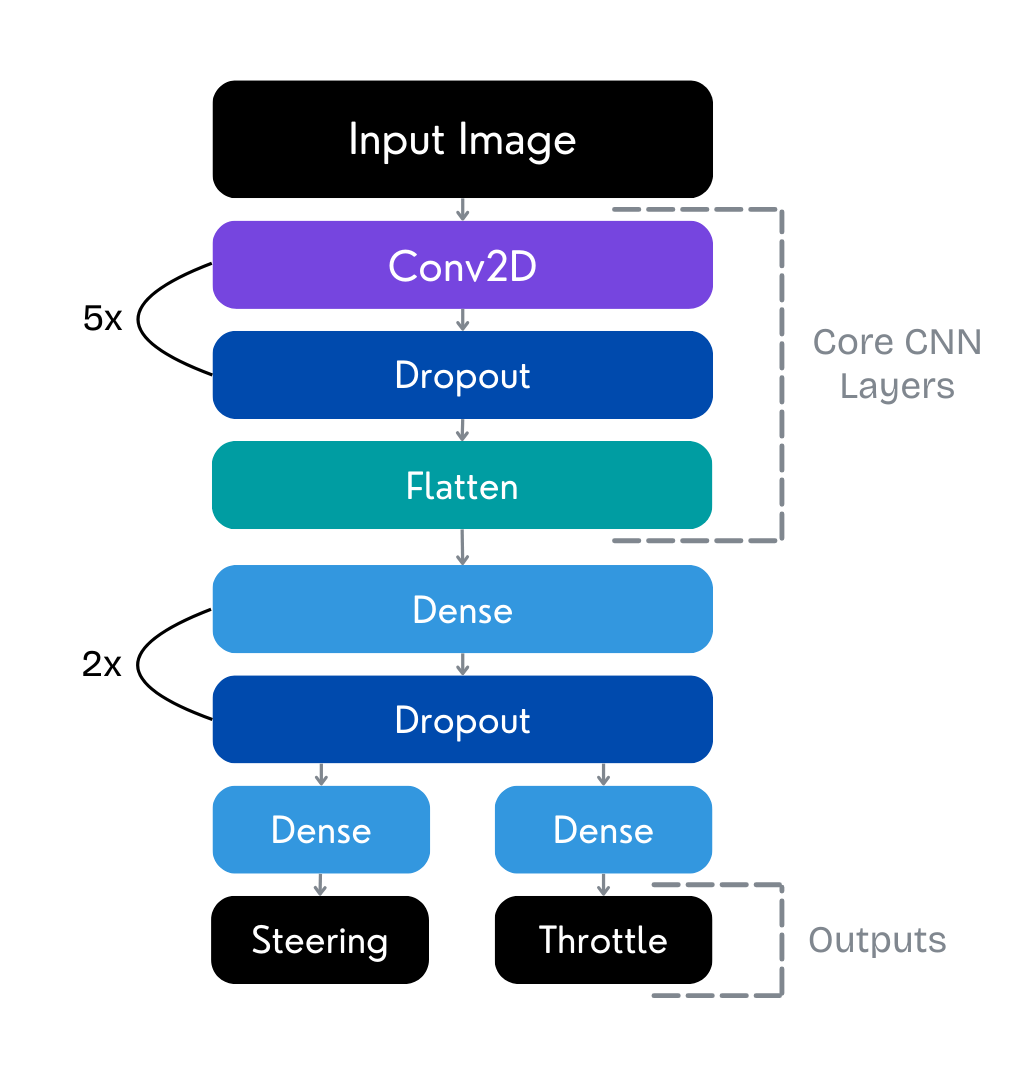}
    \caption{Network Structure.} 
    \label{fig:red}
\end{figure}

We developed two deep learning models using Keras to interpret this visual data: a Linear Convolutional Neural Network (CNN) and a Recurrent Neural Network (RNN) architecture. The CNN model, shown in Figure \ref{fig:red}, consists of 5 convolutional layers followed by two dense layers before output, and the output has two thick layers with one scalar output each, with linear activation for steering and throttle. The RNN model incorporates four time-distributed convolutional layers, followed by 2 LSTM layers, three dense layers, and driving controls; in the output, there is a thick layer with two scalar outputs for steering and throttle.

Our training dataset comprises over 10000 images and their corresponding control commands, collected during multiple runs on the test track. Images were preprocessed and resized to 160x120 pixels to optimize model performance and reduce computational load. Both models were trained up to 60 epochs.

During inference, the Jetson Nano runs the best model; the resulting commands for throttle and steering are then converted into PWM signals through the PCA9685 controller, directly interfacing with the ESC and steering servo motor.

We implemented a web-based user interface powered by Flask, enabling remote monitoring and control of the vehicle. This interface displays live telemetry, including speed, steering angle, and real-time video feed from the camera. Additionally, all sensor data, control commands, and system states are logged for offline analysis and model improvement, utilizing Pandas and Matplotlib libraries.

Overall, this software system achieves efficient and accurate autonomous navigation by combining spatial and temporal visual analysis with robust control integration from the framework, meeting the requirements of the FIRA 2025 Autonomous Cars challenge.

\section{Results}

\begin{figure}[H]
    \centering
           \includegraphics[scale=0.24]{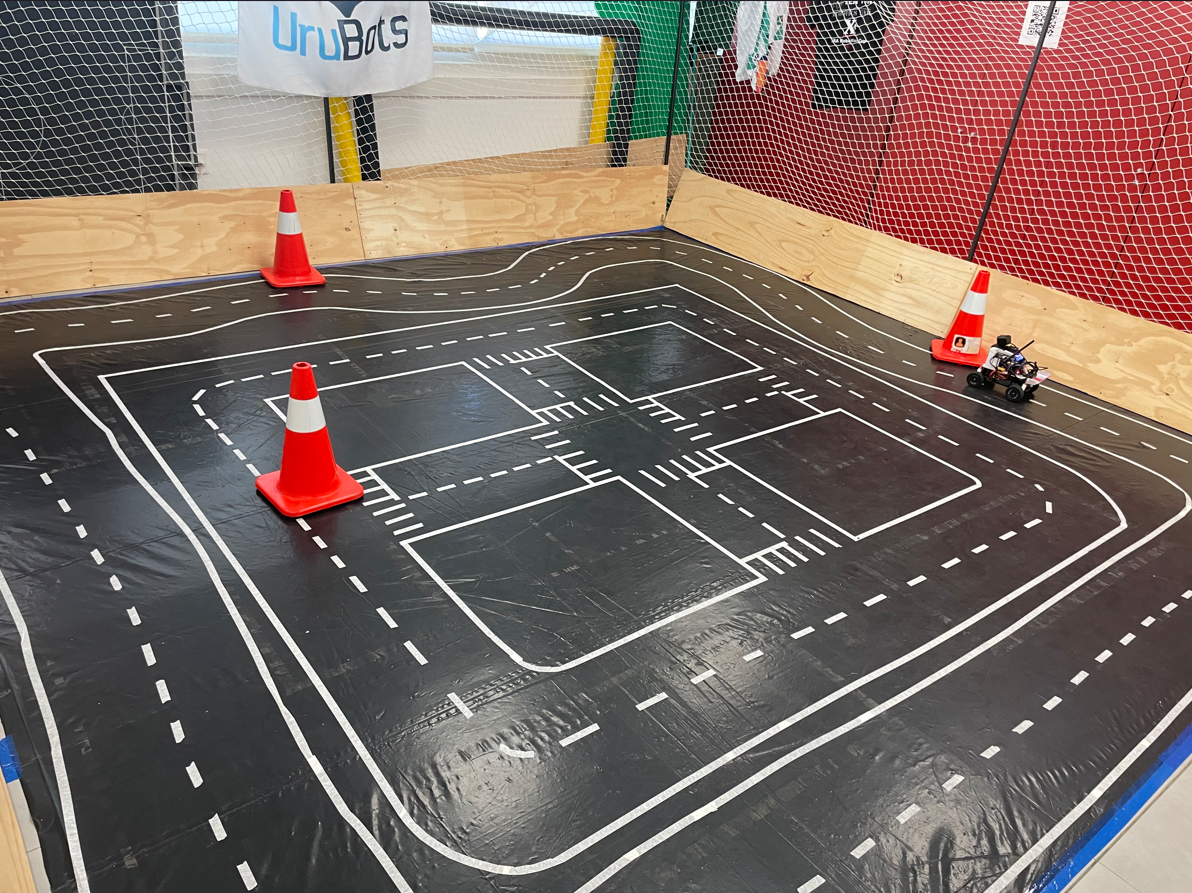}
    \caption{Scenario used to validate our vehicle} 
    \label{fig:scenario}
\end{figure}

To evaluate the performance of the autonomous vehicle, a test track of approximately 12 meters in length was designed within a 5x5 meter area, as shown in Figure \ref{fig:scenario}. Multiple runs were conducted using different trained models (CNN and RNN) in order to compare their accuracy, stability, and navigation speed.

During the tests, the vehicle was able to complete the track in an average time of 28.4 seconds, maintaining a steady speed close to 0.42 meters per second. Under normal conditions without obstacles, the vehicle followed the course with consistent accuracy. When obstacles were introduced, the RNN model showed better performance than the CNN model, demonstrating more stable and reliable decision-making, likely due to its ability to take into account the temporal sequence of frames.

Braking and reverse movement tests were also conducted to simulate the second stage of the FIRA challenge. These tests focused on evaluating the precision of low-speed control, where the new DSPower 540 45T brushed motor enabled smooth and controlled maneuvers, particularly during reverse operations and when stopping near target positions.

\begin{figure}[H]
    \centering
    \subfloat[First-person view (FPV) perspective of the car navigating the track.\label{fig:fpv}]{
        \includegraphics[width=0.45\linewidth]{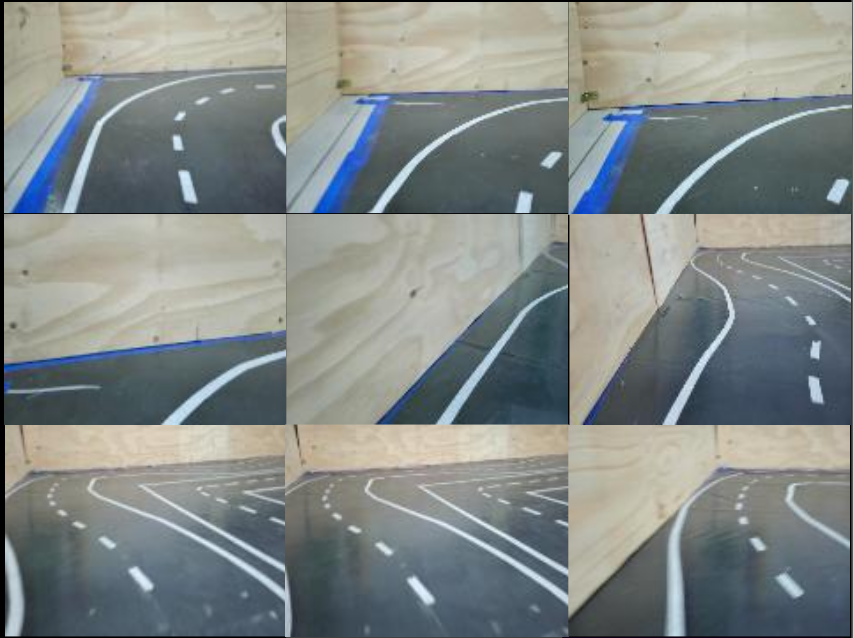}
    }
    \hfill
    \subfloat[Neural network activation or prediction visualization based on input image.\label{fig:nn_view}]{
        \includegraphics[width=0.45\linewidth]{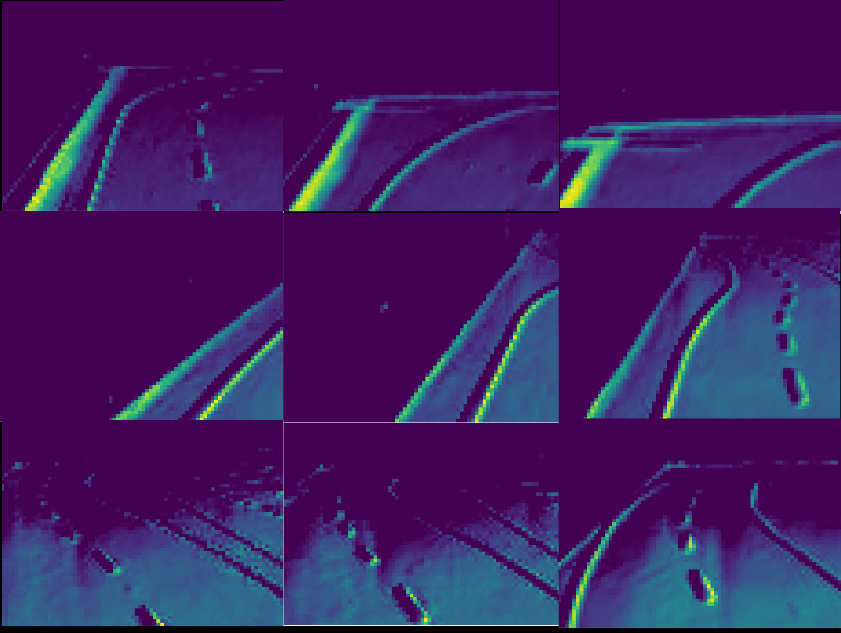}
    }
    \caption{Comparison between the actual camera input during driving (left) and the model's learned perception (right).}
    \label{fig:timeframe}
\end{figure}

To better understand the model’s behavior and perception, we include a visual representation of the system during a test run. Figure~\ref{fig:timeframe} shows a side-by-side comparison of the FPV image captured by the camera and the corresponding output or feature activation from the neural network at that moment. This frame illustrates the learned attention and decision-making pattern of the trained model.

\section{Conclusion}
In general, we conclude that our autonomous vehicle is capable of navigating reliably on test tracks, successfully completing laps and avoiding obstacles. The use of learning-based models, along with real-time image processing and stable low-speed control, has allowed the system to perform well under conditions similar to those expected in the FIRA 2025 challenge. We believe that the design we achieved, combining hardware modifications with AI techniques, confirms the ability of the model to extract relevant features from the visual input and translate them into meaningful control actions, and prepares us well to face the upcoming competition.

Additionally, the Donkey Car framework has been a great tool in supporting our development process, and we continue to build on top of it to focus specifically on the challenges of our competition category.

\bibliographystyle{splncs04}

\bibliography{sample}

\begin{thebibliography}{1}
\providecommand{\url}[1]{\texttt{#1}}
\providecommand{\urlprefix}{URL }
\providecommand{\doi}[1]{https://doi.org/#1}

\bibitem{moraes2024urubots}
Moraes, P., Peters, C., Da~Rosa, A., Melgar, V., Nu{\~n}ez, F., Retamar, M., Moraes, W., Saravia, V., Sodre, H., Barcelona, S., et~al.: Urubots autonomous cars team one description paper for fira 2024. arXiv preprint arXiv:2406.08745  (2024)

\bibitem{moraes2024behavior}
Moraes, P., Peters, C., Sodre, H., Moraes, W., Barcelona, S., Deniz, J., Castelli, V., Guterres, B., Grando, R.: Behavior cloning for mini autonomous car path following. In: 2024 IEEE URUCON. pp.~1--5. IEEE (2024)

\bibitem{pehlivan2020real}
Pehlivan, B., Kahraman, C., Kurtel, D., Nakіp, M., G{\"u}zeli{\c{s}}, C.: Real-time implementation of mini autonomous car based on mobilenet-single shot detector. In: 2020 Innovations in Intelligent Systems and Applications Conference (ASYU). pp.~1--6. IEEE (2020)

\end{thebibliography}

\end{document}